\documentclass{article} 
\usepackage{iclr2025_conference,times}

\usepackage{amsmath,amsfonts,bm}

\def\eqref#1{equation~\ref{#1}}

\def\1{\bm{1}}

\DeclareMathAlphabet{\mathsfit}{\encodingdefault}{\sfdefault}{m}{sl}
\SetMathAlphabet{\mathsfit}{bold}{\encodingdefault}{\sfdefault}{bx}{n}

\usepackage{hyperref}
\usepackage{url}

\usepackage{wrapfig}
\usepackage[table,xcdraw]{xcolor}
\usepackage{graphicx} 
\usepackage{booktabs}
\usepackage{longtable}
\usepackage{microtype}
\usepackage{subfig}
\usepackage[section]{placeins}
\usepackage{comment}
\usepackage{multirow}
\usepackage{caption}
\usepackage{relsize}
\usepackage{subcaption}
\usepackage{diagbox}
\usepackage{makecell}
\usepackage{collcell}
\usepackage{pgf}
\usepackage[dvipsnames]{xcolor}

\title{Confidence as a Reward: Transforming LLMs into Reward Models}

\author{
\textbf{He Du}\textsuperscript{\rm 1 \rm 2}\quad
\textbf{Bowen Li}\textsuperscript{\rm 2}\thanks{Corresponding author.} \quad
\textbf{Chengxing Xie}\textsuperscript{\rm 2 \rm 3}\quad
\textbf{Chang Gao}\textsuperscript{\rm 4}\quad
\textbf{Kai Chen}\textsuperscript{\rm 2}\footnotemark[1] \quad
\textbf{Dacheng Tao}\textsuperscript{\rm 5} \vspace{0.1cm}
\\
\textsuperscript{\rm 1}Fudan University \quad
\textsuperscript{\rm 2}Shanghai AI Laboratory \quad 
\textsuperscript{\rm 3}Xidian University \\
\textsuperscript{\rm 4}The Chinese University of Hong Kong \quad
\textsuperscript{\rm 5}Nanyang Technological University \vspace{0.1cm}
\\
    \texttt{elyndendu@gmail.com}~,~\texttt{libowen.ne@gmail.com}~,~\texttt{chenkai@pjlab.org.cn}
}

\newcommand{\crew}{CRew}
\newcommand{\crewdpo}{CRew-DPO}
\iclrfinalcopy 
\begin{document}

\maketitle

\begin{abstract}
Reward models enhance the reasoning capabilities of large language models (LLMs) but typically require extensive curated data and costly training, often bringing challenges.
LLM-as-a-Judge offers a training-free alternative by using LLMs’ intrinsic reasoning to evaluate responses, but it still lags behind trained reward models.
In this work, we propose \textbf{C}onfidence-as-a-\textbf{Rew}ard (\textbf{\crew}), a simple yet effective training-free approach that uses a model’s token-level confidence in the final answer as a reward proxy, particularly for close-ended problems.
Additionally, we introduce \textbf{\crewdpo}, a training method that constructs preference data from confidence scores and correctness signals. 
We validate our approach through extensive experiments on mathematical tasks. 
\crew~achieves the best performance among training-free reward models on MATH500 and RewardMATH, even surpassing most trained reward models, highlighting its effectiveness as a reward proxy. 
Additionally, we show a strong correlation between \crew~evaluations and model reasoning performance. 
Furthermore, \crew~can be used as a data filtering strategy by selecting high-quality training samples. 
Finetuning with \crewdpo~further enhances judging capabilities and outperforms existing self-training methods.
\end{abstract}

\section{Introduction}

Large language models (LLMs) have demonstrated remarkable performance on complex reasoning tasks~\citep{guo2025deepseek, gpt_o1}, significantly advancing fields such as mathematical problem-solving, logical inference~\citep{math500, tafjord2020proofwriter}, and decision-making systems~\citep{gao2023strategyllm}.
This progress has sparked growing interest in reward modeling, a critical component in refining LLM capabilities~\citep{huang2023large, xia2024evaluating, setlur2024rewarding}.
Reward models evaluate the quality of model-generated responses by typically determining their preference over baseline solutions, playing a vital role in reinforcement learning based training~\citep{schulman2017proximal, ouyang2022training}.
Beyond optimizing policy updates, reward models also enable test-time scaling by selecting the most promising solution among multiple generated candidates, thereby enhancing overall model reliability and effectiveness~\citep{hosseini2024v, snell2024scaling}.

Despite their utility, training reward models presents several fundamental challenges~\citep{rewardmath, lambert2024rewardbench}.
First, developing a well-calibrated reward model typically requires extensive curated datasets, which can be costly and time-consuming to construct~\citep{math500, math-shepherd, xia2024evaluating}. 
Second, training these models is computationally expensive and challenging, as it requires the reward model to learn subtle reasoning patterns
~\citep{yuan2024advancing, zhang2024rest}. 
Third, trained reward models may overfit to specific datasets, capturing surface-level heuristics rather than true judging capabilities, failing to generalize well beyond their training distribution~\citep{gao2023scaling}.
An alternative approach, LLM-as-a-Judge~\citep{li2024llms, gu2024survey}, leverages LLMs’ intrinsic generation capabilities to evaluate responses without explicit reward model training. 
However, existing study suggests that this method still underperforms compared to trained reward models~\citep{generative_verifier}.

\begin{figure*}[tbp!]
\vskip -0.1in
\begin{center}
\centerline{\includegraphics[width=\textwidth]{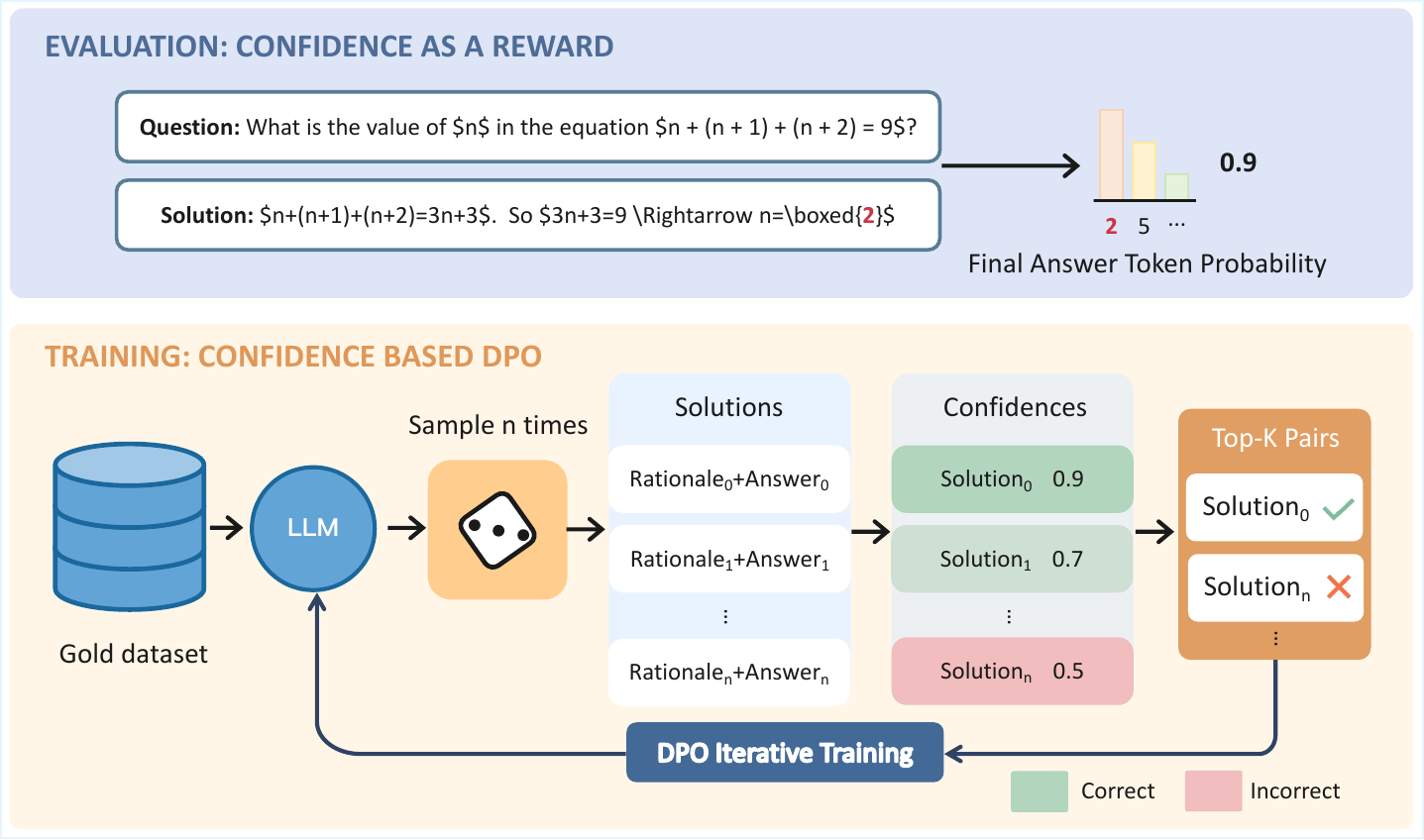}}
\caption{\textbf{An Overview of Our Approach.} The upper part illustrates how to calculate the confidence reward of a given solution. The bottom part describes how training on close-end tasks can lead to the self-improvement of the model’s evaluation ability. For a given question, the model outputs multiple solutions, each solution consists of a rationale and an answer. The answers are extracted from all the solutions to calculate the confidence. Finally, the top-K pairs with the largest confidence difference among all correct and incorrect solution pairs are selected for DPO training.}
\label{main_figure}
\end{center}
\vskip -0.3in
\end{figure*}

This raises a research question: \textit{Can LLMs themselves serve as reward models in a training-free manner, and how well can they perform?}
In this work, we introduce \textbf{C}onfidence-as-a-\textbf{Rew}ard (\textbf{\crew}), a simple yet effective approach that utilizes the model’s token-level confidence of the final answer as a reward proxy, primarily for close-ended problems where the final answer can be effortless identified and objectively verified. 
As illustrated in the upper part of Figure~\ref{main_figure}, instead of relying on extensive datasets and additional training, \crew~directly extracts the probability of the final answer tokens and computes their mean confidence score, harnessing the model’s inherent reasoning capabilities to assess response quality.
Beyond the reward estimation, we propose a corresponding training method, \textbf{\crewdpo}, to further enhance the model’s judging capabilities. 
As shown in the lower part of Figure~\ref{main_figure},
\crewdpo~constructs preference data by sampling data from the model itself and leveraging both confidence scores and solution correctness.
This preference data is then used for DPO training, enabling the model to refine its reward function without requiring external annotations.

We conduct comprehensive experiments on mathematical reasoning tasks, as they are among the most important and representative benchmarks.
Using the same base model, Qwen2.5-7B-Instruct, \crew~achieves the best performance among training-free reward modeling approaches on MATH500 \cite{math500} when applied with reward-weighted self-consistency and Best-of-N selection.
It also significantly outperforms those methods on the RewardMATH benchmark \cite{rewardmath}, demonstrating its effectiveness as a reward proxy.
Scaling to larger model variants (14B, 32B, and 72B), perhaps surprisingly, \crew~remains among the top-performing approaches on RewardMATH across all model types, even surpassing most reward models trained on massive datasets.
Additionally, we observe a strong positive correlation between a model’s mathematical reasoning performance on MATH and its evaluation capability as measured by \crew~on RewardMATH, a finding we validate across the Llama-3 and Qwen2.5 model families.
Moreover, \crew~can serve as a data filtering strategy, significantly improving model finetuning by selecting high-quality training samples.
Beyond training-free evaluation, we finetune Qwen2.5-7B-Instruct using \crewdpo.
\crewdpo~not only outperforms existing self-training methods—including vanilla DPO, ReST$^\text{EM}$ \citep{rest_em}, and LLM-as-a-Judge on MATH500 and GSM8K, but also achieves significantly higher gains on RewardMATH, further validating the effectiveness of confidence-based self-supervised learning.

Our main contributions are as follows:

\text{~~~~~~• }We introduce \crew, a novel confidence-based reward model for training-free solution evaluation, along with \crewdpo, a corresponding training approach.

\text{~~~~~~• }We demonstrate that \crew~serves as an effective reward proxy for mathematical reasoning tasks, showing a strong correlation with model performance, and utility for data filtering.

\text{~~~~~~• }We validate the effectiveness of \crewdpo~through training experiments, showing significant improvements in evaluation capability over existing self-training methods.

\label{sec:intro}

\section{Related Work}

\textbf{ORM and PRM}. The output generated by ORM based models typically focuses solely on the final result when calculating rewards, without paying attention to the finer details of the process~\citep{yuan2024advancing, cai2024internlm2, liu2024skywork,dai2023safe, yang2024regularizing, ArmoRM}. Generally, pairs consisting of correct and incorrect reasoning data are constructed and used for training with pairwise loss~\citep{bradley1952btmodel}. Our method essentially transforms a mathematical model into an ORM with the confidence format.
~\citet{math500} suggests that PRM focuses on every step of the reasoning process and assigns a reward value to each step. Training often requires a large amount of manually labeled, fine-grained supervisory data for each step~\citep{xia2024evaluating, skyworkopeno12024}. To address the heavy reliance on manually labeled data, ~\citet{math-shepherd}, ~\citet{zhang2024rest} and other works mimic reinforcement learning by defining the reasoning process as a Markov Decision Process (MDP). This approach allows the use of multiple rollout data to train PRM~\citep{xie2024monte, feng2023alphazero}.

\textbf{Generative Verifier}. ~\citet{generative_verifier} uses the probability of a ``Yes" or ``No" token indicating the correctness of a solution as the reward value. This approach leverages the text generation capabilities of pretrained large language models (LLMs) through the CoT (Chain-of-Thought) in the critic process~\citep{wei2022chain}. ~\citet{wang2024chain} focuses on the process of sampling tokens, referring to the difference in probability between the top-1 and top-2 choices in the final answer part as confidence. It argues that the greater this confidence, the more it indicates that the model has arrived at the answer through a series of reasoning processes. Our method differs from the above methods by directly using the probability of the final answer in the solution as the reward. This allows us to obtain fine-grained rewards while maintaining consistency with the reasoning process. For the model, it can directly obtain the reward after reasoning, without the need for an additional calculation process.

\textbf{LLM-as-a-Judge}. ``LLM-as-a-Judge" is not actually a single method but rather a broad concept. In a wider sense, any approach that utilizes the inherent attributes and capabilities of LLMs falls under this category, mainly based on prompt-based methods~\citep{fu2023gptscore, wei2022chain, dong2024can, lin2023llm, bai2022constitutional, Ling_Fang_Li_Huang_Lee_Memisevic_Su_Diego_Research, zheng2023judging}. It relies on the model’s judgment ability, which is gained through large-scale pretraining. ~\citet{self_rewarding} demonstrates that such rewards can, to some extent, reflect the quality of reasoning. However, due to its training-free nature, it may encounter issues of instability and difficulty in handling complex tasks~\citep{li2024llms, gu2024survey}.

\section{Methodology}
\label{theory}
\subsection{\crew}
Existing reward methods—both discriminative and generative—often directly evaluate the entire solution, which can bring several issues. For example, learning to evaluate the whole solution is relatively difficult and usually requires more training. Additionally, key tokens within the solution may be obscured, resulting in suboptimal reward calculations.

To address the mentioned issue, our method introduces two improvements:

\text{~~~~~~• }Focus on the final answer tokens: We only consider the key tokens corresponding to the final answer, providing a more accurate reward.

\text{~~~~~~• }Align reward computation with the model's reasoning process: The reward calculation process is identical to the model's reasoning process, allowing for better utilization of the model's capabilities, which have been honed through extensive training.

\paragraph{What is confidence?} 
As shown in the upper part of Figure~\ref{main_figure}, to capture the most important information from the solution, we focus on extracting tokens corresponding to final answers (e.g. content within \texttt{\textbackslash boxed\{\}} in MATH tasks) and compute their mean probability as confidence. This confidence metric leverages the model's inherent reasoning capabilities through Chain-of-Thought generation, effectively transforming its reasoning process into a quantifiable reward signal that reflects solution certainty.

\paragraph{Why confidence can be used as a reward?}
Let’s start by defining an existing model $\pi$, and for a given question $q$ with gold solution $s$ containing the gold answer $a$. 
This model generated solution $\hat{s}$ consists of two parts: the rationale $\hat{r}$ and the final answer $\hat{a}$, so that $\hat{s} = <\hat{r}, \hat{a}>$.
At the same time, we define a verification function $v$, where $v(\hat{s},q)=1$ if the solution is correct, and $v(\hat{s}, q)=0$ otherwise.

Close-ended reasoning tasks' characteristic implies that the probability of a solution being correct is equal to the probability that the final answer is the gold answer:
\begin{equation}
    P(v(\hat{s}, q)=1) = P(\hat{a}=a)
\end{equation}

However, the equivalence between a correct solution and the answer being the gold answer is not sufficient to demonstrate that confidence is an excellent reward. Therefore, we will next explain the conditions that a reward needs to satisfy to be considered effective.

We define a reward function \( R \) as a binary classification function, where ideally for a solution \( s \), if \( s \) is correct, \( R(s) = 1 \); otherwise, \( R(s) = 0 \). To approach this goal, one way is to ensure that at least the following conditions are met:

\textit{For question $q$, suppose we have two reward functions, $R_1$ and $R_2$, where $R_2$ is the better one, and define a set 
$S = \{S^{+}, S^{-}\}$ that contains all possible solutions. For all possible correct solutions \( s^{+} \in S^+\)  and incorrect solutions 
\( s^{-} \in S^- \), we require:}
\begin{align}
\label{eq:req}
  \mathbb{E}_{s^{+} \in S^{+}}[R_{2}(s^{+}) - R_1(s^{+})] &> 0  \notag \\
  \mathbb{E}_{s^{-} \in S^-}[R_{2}(s^{-}) - R_1(s^{-})] &< 0
\end{align}
At the same time, we present a hypothesis involving two models, $\pi_1$ and $\pi_2$ , where 
$\pi_2$ has stronger reasoning capabilities than 
$\pi_1$ . The hypothesis is as follows:
\begin{align}
\label{eq:hypothesis}
    \mathbb{E}_{\hat{s}_2\sim \pi_2}[P_{\pi_2}(v(\hat{s}_2, q)=1)] &> \mathbb{E}_{\hat{s}_1\sim \pi_1}[P_{\pi_1}(v(\hat{s}_1, q)=1)] \notag \\
    \mathbb{E}_{\hat{s}_2\sim \pi_2}[P_{\pi_2}(v(\hat{s}_2, q)=0)] &< \mathbb{E}_{\hat{s}_1\sim \pi_1}[P_{\pi_1}(v(\hat{s}_1, q)=0)]
\end{align}

Therefore, by simply substituting confidence into Hypothesis \ref{eq:hypothesis}, we can easily derive Requirement \ref{eq:req}, which means that confidence can be a form of ideal reward function (The complete derivation can be found in the \ref{app::Derivation}). 
This also demonstrates that, there is a correlation between the model's reasoning ability and its evaluation capability.

\subsection{\crewdpo}
Based on confidence, the new reward calculation method, we propose a corresponding training approach, \crewdpo. \crewdpo~ is a self-training approach that does not require a large amount of manually labeled data, yet significantly improves confidence performance.

\textbf{Data Preparation.}
As shown in the bottom part of Figure~\ref{main_figure}, given a seed dataset consisting of a series of questions with gold solutions. We first define the current model as $\pi_{t}$, and the seed dataset with gold answers as $D = \{(q_i, s_i)\}$. The current model samples $N$ different solutions for each question and computes the confidence for each solution:
\[
c_i^n = \pi_t(a_i^n|r_i^n, q_i) \quad  n \in \{1,2,\dots,N\},
\]
Finally we get a new dataset:
\[
    \hat{D} = \{(q_i, \{s_i^n, c_i^n\})\} \quad n \in \{1,2,\dots,N\}.
\]
The solutions generated for all questions in $\hat{D}$ are classified into two sets, $\hat{D}^{chosen}$ and $\hat{D}^{rejected}$ , based on whether they match the gold answer. Specifically, 
$\hat{D}^{chosen}$ contains solutions that are correct, and $\hat{D}^{rejected}$ contains solutions that are incorrect. Then, for each question, the solutions from $\hat{D}^{chosen}$ and $\hat{D}^{rejected}$ are matched, forming a set that contains all possible correct-incorrect solution pairs:

\begin{align}
    \hat{D}^{pairs} = \{(q_i, \{s_i^{m},c_i^{m}\}_{chosen},\{s_i^{n},c_i^{n}\}_{rejected})\}, \quad m+n \in \{2,\dots,N\}. \notag
\end{align}

Once $\hat{D}^{pairs}$ is obtained, we select the top-K pairs with the largest confidence differences between the correct and incorrect solutions. These pairs are then used to form the training set $\hat{D}^{train}$ for subsequent preference optimization.

\textbf{Preference Optimization.} With the constructed pair training set $\hat{D}^{train}$, the next stage is to improve the model's evaluation capability through preference optimization.  We update the initial model on $\hat{D}^{train}$ using DPO loss~\citep{rafailov2024direct}, which results in the new model.

The reason for this training approach is to enable the model to better associate the correctness of the solution with the level of confidence. Specifically, a correct solution should correspond to a higher confidence, while an incorrect solution should have a lower confidence.

\section{Experiments with \crew}

In this section, our experiments focus on the superiority of using confidence as a reward, 
including its excellent evaluation performance, strong correlation with the model's inherent reasoning capabilities and its ability to help filter training data.

\subsection{Setup}
\paragraph{Datasets}
For the most important and representative reasoning task—mathematics, we focus on two widely used datasets: the popular grade-school math test dataset GSM8K~\citep{gsm8k} and the more challenging MATH dataset~\citep{math}. 
For evaluating mathematical reasoning tasks on MATH, we select the representative MATH500 dataset~\citep{math500} from the same source to achieve higher evaluation efficiency. 
Specifically, when evaluating the performance of the reward model, in addition to using indirect methods like majority voting based on rewards, we also use a direct dataset called RewardMATH~\citep{rewardmath}. 
This dataset is built on MATH500 and includes 1 correct solution and 9 incorrect solutions for each question, randomly generated by 14 popular open-source and closed-source models. 
The objective of RewardMATH is to select the correct solution from a total of 10 solutions (for example, please refer to the Table~\ref{rewardmath_example}), which directly reflects the reward model's ability to assess mathematical reasoning solutions.

\subsection{\crew~as a Strong Reward Proxy}

One of the advantages of confidence is that it is a training-free reward. Training-free reward means that for a solution generated by a model, there is no need to train an additional reward model to obtain the reward value for that solution.

In this section, we compare confidence with several different training-free reward methods on MATH500 and RewardMATH datasets, such as LLM-as-a-Judge, generative verifier, and perplexity.
For LLM-as-a-Judge, we follow the prompt from ~\citet{self_rewarding}, asking the model to score its own output. 
For the generative verifier method~\citep{generative_verifier}, after the model generates a solution, we ask the same model (which has not been trained on specific evaluation tasks) whether the answer is correct, and then uses the probability of the Yes or No token in the model's response as the reward. 
Similar to the calculation of confidence, perplexity—one of the basic properties of model output—can also be used as a form of reward.

The two parts below Table~\ref{tab:RewardMATH_results} show the performance of all training-free reward methods on RewardMATH. As seen, confidence outperforms the other methods, highlighting its ability to distinguish between correct and incorrect solutions. Table~\ref{self_reward_results} presents the performance of the same model on MATH500 using SC+Reward(where the normalized reward is used as the weight for voting) and best-of-N approaches, under different training-free reward approaches. The result also shows that confidence achieves the best performance.

After our analysis, the LLM-as-a-Judge method, tends to assign higher scores during evaluation, which impacts the evaluation process negatively.
Additionally, due to the integer scoring form, it cannot distinguish between solutions with the same score, resulting in lower evaluation accuracy. 
For the generative verifier method, it reduces the granularity of the reward, but since evaluating solutions has not been extensively trained for the model, its capability remains relatively weak. 
As for perplexity, compared to confidence, experiments show that the perplexity reward may include the probability of many irrelevant tokens, which leads to a less accurate evaluation of the solution.


It is evident that, whether on MATH500 or RewardMATH, confidence proves to be a more effective training-free reward, distinguishing between correct and incorrect solutions more effectively. At the same time, when evaluating the content generated by the model itself, there is no need for the overhead of an additional reward calculation process.

\begin{wraptable}{r}{0.5\textwidth}
\vskip -0.1in
\centering
\setlength{\tabcolsep}{10pt}  
\resizebox{!}{5.6cm}{
\setlength{\tabcolsep}{10pt}  
\begin{tabular}{lcc}
\toprule

\textbf{Reward Model} & & \textbf{RewardMATH} \\
\midrule
\textit{Random} & & 10.00 \\

\midrule
\multicolumn{3}{c}{\cellcolor{gray!11}\textbf{\textit{Trained Classifier-based Reward Models}}} \\
\href{https://huggingface.co/Ray2333/GRM-Gemma-2B-sftreg}{GRM-gemma-2B} & & 4.97 \\
\textbf{Qwen2.5-7B-Instruct-rm$^\$$} & & 6.83 \\
\href{https://huggingface.co/OpenAssistant/oasst-rm-2.1-pythia-1.4b-epoch-2.5}{Oasst-rm-2.1-pythia-1.4b} & & 7.04 \\
\href{https://huggingface.co/PKU-Alignment/beaver-7b-v2.0-reward}{Beaver-7b-v2.0-reward} & & 7.25 \\ 
\href{https://huggingface.co/openbmb/Eurus-RM-7b}{Eurus-RM-7b} & & 16.98 \\ 
\href{https://huggingface.co/RLHFlow/ArmoRM-Llama3-8B-v0.1}{ArmoRM-Llama3-8B-v0.1} & & 20.50 \\
\href{https://huggingface.co/Skywork/Skywork-Reward-Llama-3.1-8B}{Skywork-Reward-Llama3.1-8B} & & 22.15 \\
\href{https://huggingface.co/Ray2333/GRM-llama3-8B-sftreg}{GRM-llama3-8B} & & 24.43 \\
\href{https://huggingface.co/internlm/internlm2-20b-reward}{Internlm2-20b-reward} & & 33.95 \\
\href{https://huggingface.co/internlm/internlm2-7b-reward}{Internlm2-7b-reward} & & 37.27 \\ 
\midrule
\multicolumn{3}{c}{\cellcolor{gray!11}\textit{\textbf{Trained Process Reward Models} (prod)}} \\
\href{https://huggingface.co/ScalableMath/llemma-7b-prm-prm800k-level-1to3-hf}{Llemma-7b-prm-prm800k} & & 14.08   \\
\href{https://huggingface.co/GAIR/ReasonEval-34B}{ReasonEval-34B} & & 15.95   \\
\href{https://huggingface.co/peiyi9979/math-shepherd-mistral-7b-prm}{Math-Shepherd-Mistral-7B} & & 17.18 \\
\href{https://huggingface.co/GAIR/ReasonEval-7B}{ReasonEval-7B} & & 18.22   \\ 
\multicolumn{3}{c}{\cellcolor{gray!11}\textit{\textbf{Trained Process Reward Models} (geo mean)}} \\
\href{https://huggingface.co/peiyi9979/math-shepherd-mistral-7b-prm}{Math-Shepherd-Mistral-7B} & & 15.74 \\
\href{https://huggingface.co/ScalableMath/llemma-7b-prm-prm800k-level-1to3-hf}{Llemma-7b-prm-prm800k} & & 16.36   \\
\href{https://huggingface.co/GAIR/ReasonEval-34B}{ReasonEval-34B} & & 18.43   \\
\href{https://huggingface.co/GAIR/ReasonEval-7B}{ReasonEval-7B} & & 20.29   \\ 
\midrule
\multicolumn{3}{c}{\cellcolor{gray!11}\textit{\textbf{Training-Free Reward}}} \\
Perplexity & & 2.07 \\
LLM-as-a-Judge & & 13.25 \\
Generative Verifier & & 16.56 \\ 
\midrule
\multicolumn{3}{c}{\cellcolor{gray!11}\textit{\textbf{\crew} (Ours)}} \\
Qwen2.5-14B-Instruct & & 32.71 \\
Qwen2.5-32B-Instruct & & 40.17 \\
Qwen2.5-72B-Instruct & & 38.51 \\
\textbf{Qwen2.5-7B-Instruct} & & 27.12 \\ 
~~~~\small{\crewdpo~Iteration 1} & & 36.02   \\ 
~~~~\small{\crewdpo~Iteration 2} & & \textbf{38.72}   \\ 
\bottomrule

\end{tabular}
}
\caption{The results of our confidence reward method compared to classifier-based ORMs/PRMs reported in the ~\citet{rewardmath} and training-free reward methods (based on Qwen2.5-7B-Instruct) on RewardMATH. $\$$: The ORM trained on the same data sampled from the first iteration of CRew-DPO on MATH task, based on Qwen2.5-7B-Instruct.}
\vskip -0.5in
\label{tab:RewardMATH_results}
\end{wraptable}

\subsection{Stronger Models are Better Evaluators}
\label{sec::relation}
To validate the conclusion presented in Section \ref{theory}, that a stronger reasoning ability of the model corresponds to a stronger evaluation capability, 
in this section, we conduct correlation experiment and demonstrate a strong correlation between the two capability.

\begin{figure}[htbp!]
\begin{minipage}[c]{0.5\textwidth}
\centering
\centerline{\includegraphics[width=\textwidth]{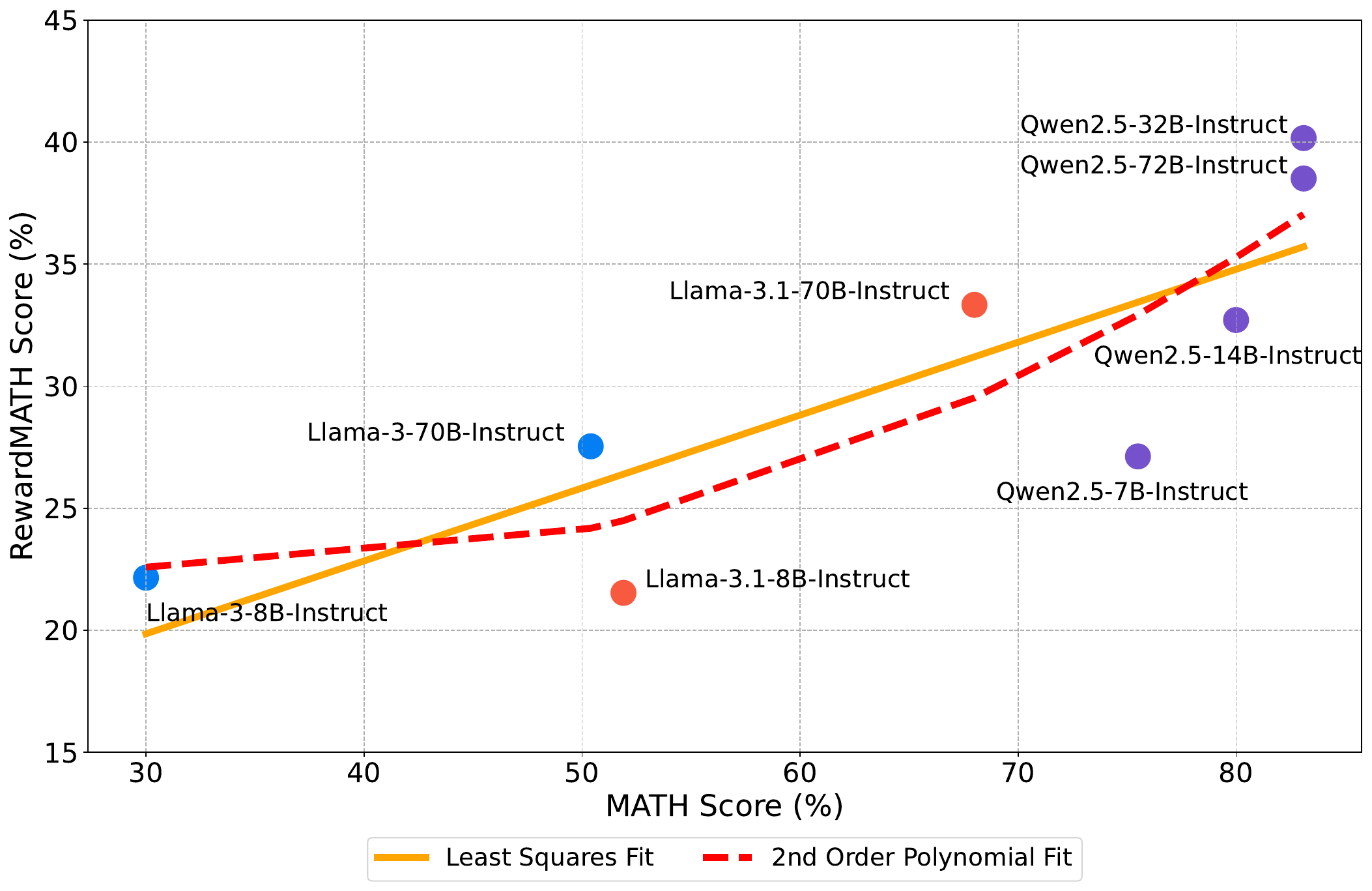}}
\caption{The correlation between mathematical reasoning ability and evaluation ability. Using eight models with varying capabilities from the Llama-3 and Qwen-2.5 families, the results demonstrate a strong positive correlation, with a correlation coefficient reaches 0.83. This demonstrates the consistency between mathematical reasoning ability and mathematical evaluation ability.}
\vskip 0.1in
\label{coefficient}
\end{minipage}
\hfill
\begin{minipage}[c]{0.45\textwidth}
\centering
\vskip -0.1in
\resizebox{!}{1.6cm}{
\begin{tabular}{lccc}
\toprule
Method                     & \multicolumn{2}{c}{MATH500}   \\ 
\cmidrule(lr){2-3}
 & SC+Reward  & BoN &     \\
\midrule
Self-consistency           & 78.2  & - \\
LLM-as-a-Judge             & 78.2  & 64.0 \\
Generative verifier        & 78.8  & 67.4 \\
Perplexity                 & 79.0  & 55.8 \\
Confidence                 & \textbf{79.2}  & \textbf{72.2} \\
\bottomrule
\end{tabular}
}
\captionof{table}{The results of different training-free reward methods on MATH500 with SC+Reward and Bon. We used Qwen2.5-7B-Instruct \citep{qwen25}. Specifically, on MATH500, we sample 16 solutions each question with temperature of 1.0, using the normalized reward as the weight for voting or choosing. All evaluations are conducted using our zero-shot prompt, instructing the model to perform step-by-step reasoning and format the final output accordingly.}
\label{self_reward_results}
\end{minipage}
  
\vskip -0.3in
\end{figure}

In the experiment, we select eight commonly used open-source models, from the Llama-3~\citep{MetaAI2024} and Qwen-2.5 families~\citep{qwen25}. 
We use the scores of all models under the official 4-shot prompt setting on the MATH testset~\citep{math} as an estimate of their mathematical capabilities and then test their performance on RewardMATH to assess their evaluation ability.
Figure~\ref{coefficient} illustrates the relationship between reasoning and evaluation. As shown, the correlation coefficient between the two reaches 0.83, indicating a strong correlation. The experiment demonstrates a strong correlation between the model's mathematical ability and evaluation capability, and also means that if we already have a model with strong reasoning capabilities, a powerful reward model can be achieved without additional training.

\subsection{\crew~as a Data Filtering Stretegy}

In this section, we aim to explore whether there is a correlation between confidence and data quality. To investigate this, we first define high-quality data as data that is more beneficial for training a weaker base model. When all other factors are kept constant, the base model trained on higher-quality data should perform better in testing.

The experimental process is as follows: for the same dataset (where a single question has multiple correct solutions), a separate confidence reward model is used to calculate the confidence for all solutions. The data with the highest and lowest confidence scores are selected, and the original gold solution is used as the baseline. The base model is then trained with the same settings using these three types of data.

We use Qwen2.5-7B-Instruct model to sample 30 solutions for each question in the MATH training set, which forms the data source required for our experiment.
To eliminate randomness, we conduct three sets of experiments using three different confidence reward models. Each set followed the same experimental process, using the same dataset and training the same base model. 

\begin{wrapfigure}{r}{0.5\textwidth}
\vskip -0.4in
\begin{center}
\centerline{\includegraphics[width=0.5\textwidth]{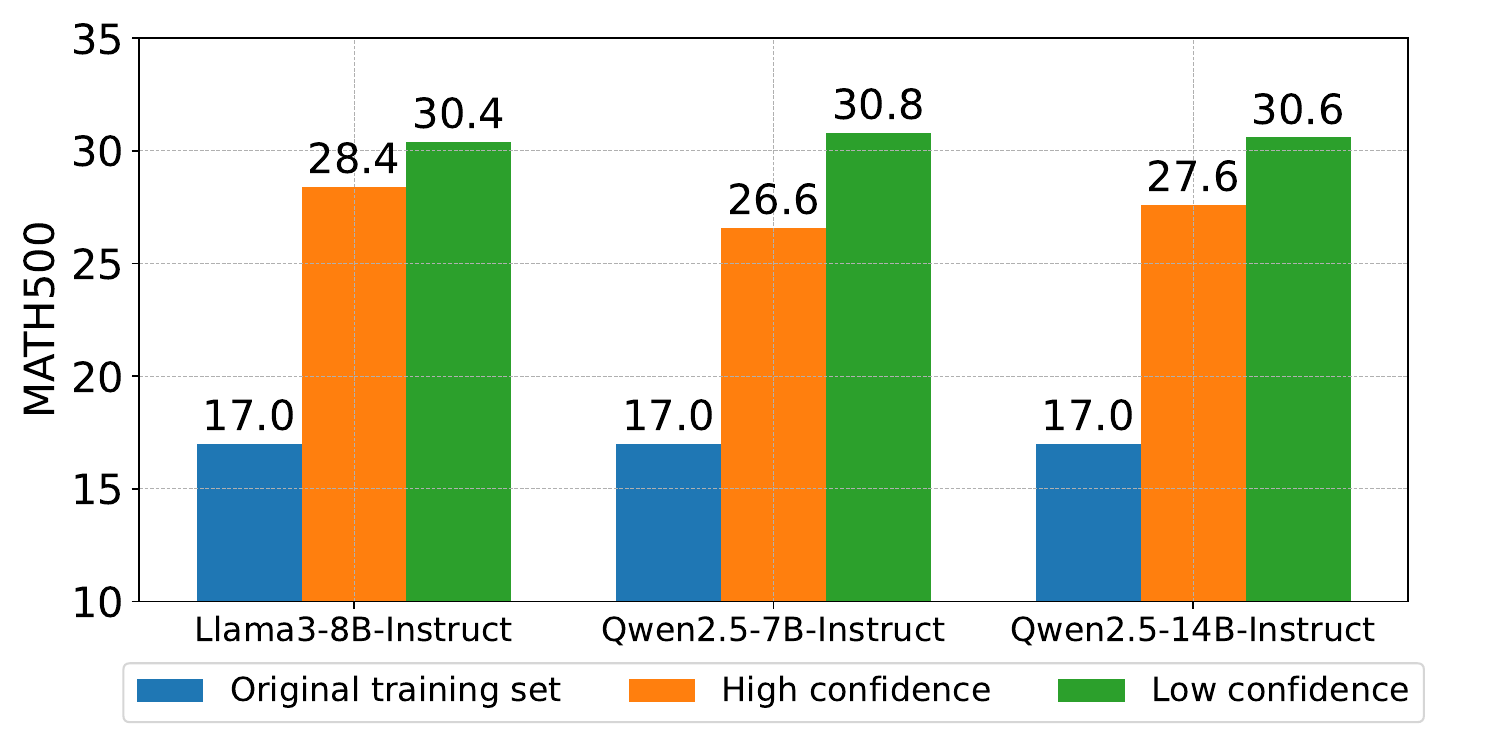}}
\caption{The results of the Llama3-8B-Base model on MATH500 using the data with the highest and lowest confidence (calculated by three different critic models.) versus the original training set, all under the same training settings.}
\label{quality}
\end{center}
\vskip -0.3in
\end{wrapfigure}

Specifically, since the dataset in the experiment was sampled by Qwen2.5-7B-Instruct, we selected an additional base model, Llama3-8B-Base, to exclude potential issues arising from similar data distributions.

As shown in Figure~\ref{quality}, the manually labeled original training set is difficult for the model to learn effectively. This supports the conclusion in \citet{setlur2024rl} that data sampled by models is easier to learn from than manually labeled data.
It is also noteworthy that models trained on low-confidence data perform significantly better than those trained on high-confidence data. Different models calculating confidence consistently show this result, indicating that solutions that are correct but where the model is less certain are more suitable for learning. These solutions may contain more potential variability, making them higher-quality data for the model.

\section{Experiments with \crewdpo}
In this section, the goal of our experiments is to validate the effectiveness of our training method, \crewdpo. \crewdpo~is a self-training approach (the model samples data to train itself) that does not rely on large amounts of externally labeled data. (The training details can be found in the Section~\ref{training_detail}.) Compared to other self-training methods, our approach not only outperforms them in terms of performance on mathematical tasks but also significantly enhances the model's evaluation capability.
\subsection{Setup}
\label{train_exp}
\paragraph{Baselines}
As comparison baselines, we compare other self-training approaches like DPO, ReST$^\text{EM}$~\citep{rest_em} and LLM-as-a-Judge~\citep{self_rewarding}:

\text{~~~~~~• }\textbf{DPO} A correct and incorrect solution pair is randomly selected from the model's sampled results, and then DPO training is applied directly.

\text{~~~~~~• }\textbf{ReST$^\text{EM}$} For each question, up to 10\footnote{
We follow the original paper's numerical settings.} correct solutions are randomly selected from the model's sampled results. During each training iteration, ReST$^\text{EM}$ performs SFT on the initial model.

\text{~~~~~~• }\textbf{LLM-as-a-Judge} The core idea is to use the model score its own sampled solutions, in order to obtain preference data, which is then used for training with DPO.

\begin{table*}
\vskip -0.3in
\begin{center}
\resizebox{!}{2.7cm}{
\begin{tabular}{l|cc|cc}
\toprule
\textbf{Task} & \multicolumn{2}{c|}{\textbf{MATH}} & \multicolumn{2}{c}{\textbf{GSM8K}}   \\
\midrule
\textbf{Dataset} & MATH500 & RewardMATH & GSM8K & RewardMATH   \\
\midrule
Qwen2.5-7B-Instruct & 71.2 & 27.12 & 90.37 & 27.12 \\
\midrule
DPO & 73.0 & 29.19 & 92.27 & 27.33 \\
\midrule
ReST$^\text{EM}$ & & & & \\
~~~~ {\em Iteration 1} & 74.8 & 28.78 & 92.42 & 26.09 \\
~~~~ {\em Iteration 2} & 72.4 & 29.81 & 91.81 & \textbf{30.43} \\
\midrule
LLM-as-a-Judge & & & & \\
~~~~ {\em Iteration 1} & 74.0 & 27.12 & 92.12 & 27.33 \\
~~~~ {\em Iteration 2} & 72.4 & 27.12 & 92.12 & 28.57 \\
\midrule
\crewdpo & & & & \\
~~~~ {\em Iteration 1} & \textbf{75.0} & 36.02 & 92.42 & \textbf{30.43} \\
~~~~ {\em Iteration 2} & 71.4 & \textbf{38.72} & \textbf{92.57} & 29.81 \\
\bottomrule
\end{tabular}
}
\end{center}
\caption{The comparison of \crewdpo~with other self-training methods in terms of evaluation ability and mathematical reasoning ability, tested separately on the MATH and GSM8K tasks, along with the corresponding RewardMATH scores.}
\label{train_result}
\end{table*}

\subsection{A Significant Improvement in Evaluation Ability} 
As shown in Table~\ref{train_result}, consistent with the conclusion validated in Section~\ref{sec::relation}, different self-training methods, while enhancing mathematical ability, have also led to performance improvements on RewardMATH to some extent. At the same time, \crewdpo~not only outperforms other baseline methods on mathematical tasks but also achieves a significantly greater improvement in evaluation ability. Across two iterations, our method improves evaluation performance by 8.9\% and 11.5\%, respectively. 

Especially in the last part of Table~\ref{tab:RewardMATH_results}, it can be seen that after training with \crewdpo, the 7B model's performance on RewardMATH reaches the level of the 72B model, demonstrating that our approach targetedly enhances evaluation capability.

Based on these observations, while ReST$^\text{EM}$ can improve mathematical task performance, the lack of erroneous data during training hinders the model's ability to learn how to evaluate a solution when the answer is unknown. At the same time, performing SFT on the instruct model after RLHF~\citep{ouyang2022training} alignment may, to some extent, degrade the evaluation capability that the model originally gained through the alignment training. For DPO and LLM-as-a-Judge,  although these methods leverage incorrect data, they do not use confidence as the selection signal. This fundamentally introduces a distribution mismatch between the data construction phase and the evaluation phase, resulting in limited performance improvements.

\subsection{The Generalization of \crewdpo}
\label{generalization_exp}

To verify whether \crewdpo~enhances evaluation capability by learning the intrinsic mathematical ability, in this experiment, we use two mathematically distinct datasets, MATH and GSM8K, with different distributions. The experiment is divided into two main components:

\textbf{Same Actor Model:} We use Qwen2.5-7B-Instruct as the actor model. The actor model will be tested on both MATH and GSM8K, guided by the reward model.

\textbf{Different reward models}, includes:

\text{~~~~~~• }Confidence reward models trained using our method on MATH or GSM8K.

\text{~~~~~~• }An untrained model (identical to the actor model) to verify whether the trained reward model improves evaluation capability compared to the untrained one.

When testing the model's evaluation capability, we continue to use the SC+Reward method described earlier.

\begin{wrapfigure}{r}{0.5\textwidth}
\vskip -0.3in
\begin{center}
\centerline{\includegraphics[width=0.5\textwidth]{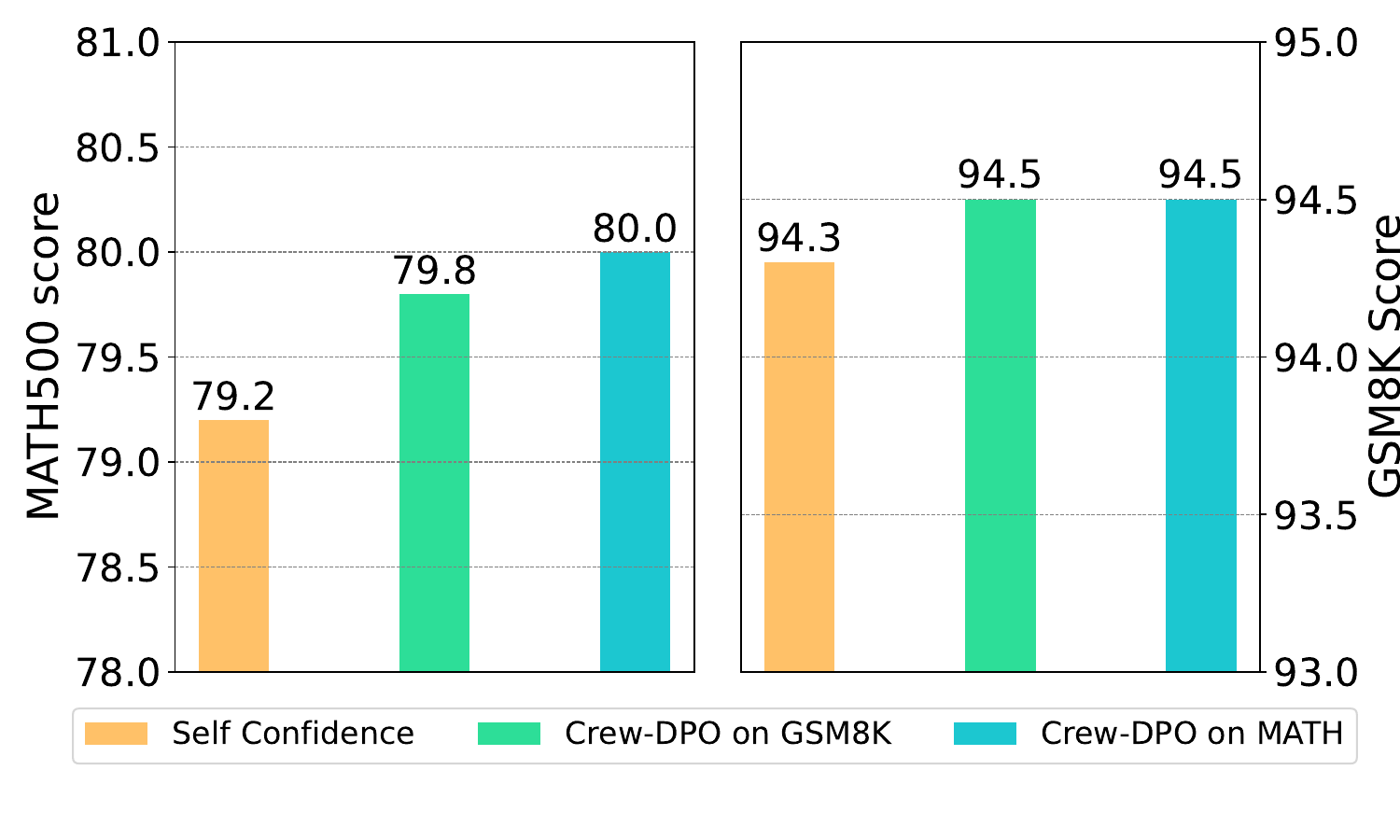}}
\caption{
A comparison between the confidence reward model trained on MATH, the confidence reward model trained on GSM8K, and the untrained model itself.
The two subplots on the left and right respectively show the performance of the actor model on MATH and GSM8K under the guidance of different reward models. The testing method here is the same as described earlier, i.e., SC+Confidence, with a temperature of 1.0 and 16 solutions sampled for each question. 
Self confidence means using the untrained actor itself to calculate the confidence.
}
\label{generalization}
\end{center}
\vskip -0.4in
\end{wrapfigure}

As shown in FIgure~\ref{generalization}, first, whether in the MATH or GSM8K tasks, using the trained confidence reward model to guide SC+Reward consistently outperforms using the actor model itself as the reward model. This demonstrates that \crewdpo~indeed leads to an improvement in evaluation capability. Furthermore, when the training task of the confidence reward model differs from the task it guides the actor's reasoning on, the confidence reward model trained with our method still brings performance improvements. This indicates that the improvement in evaluation capability brought by \crewdpo~generalizes across different mathematical tasks.

\subsection{Comparison with Trained Reward Models}

Previous experiments have already shown that models with strong mathematical abilities can also serve as excellent reward models under the confidence reward. In this section, we further demonstrate the significant enhancement of evaluation ability through confidence by comparing it with mainstream trained ORMs and PRMs on RewardMATH.

As shown in Table~\ref{tab:RewardMATH_results}, under the confidence format, even models that are not specifically trained can perform as well as the top reward models. After two iterations of training \crewdpo, the model's performance surpasses all other reward models that trained on manually labeled datasets, such as PRM800K ~\citep{math500}, or synthetic data, like ~\citet{math-shepherd}, etc. This suggests that training a reward model does not fully leverage the inherent mathematical abilities of the original model. 

Meanwhile, as shown in Table~\ref{tab:RewardMATH_results}, using the same data constructed by our training experiment on the MATH task, we trained a binary classifier ORM based on Qwen2.5-7B-Instruct. It can be observed that training a new ORM on the same data does not perform as well as the confidence-based approach. This indicates that our approach is more efficient than direct training in enhancing evaluation capabilities, and better transforms the model's inherent reasoning ability into evaluation ability.

\section{Conclusion}
In this paper, we propose a new reward method—\crew. For close-end tasks, \crew~employs the confidence of the final answer tokens within the solution as the primary reward metric. Specifically, this confidence is calculated as the mean probability of the tokens that constitute the final answer.
Beyond proposing this confidence-based reward mechanism, we also present a corresponding training method, \crewdpo, which conducts DPO training based on confidence and correctness. Experimental results show that confidence outperforms other reward methods, proving to be an effective reward. It can utilize the model's own reasoning ability and is effective for data filtering. Moreover, the training method we designed further significantly enhances the performance of confidence.

\section{Acknowledgement}

This project is supported by Shanghai Artificial Intelligence Laboratory. We are committed to advancing artificial intelligence technology through cutting-edge research and practical applications, contributing to technological progress. We are grateful for the valuable support from Shanghai AI Lab, which enables us to continue exploring new opportunities in the field of artificial intelligence and develop more advanced solutions.

\bibliography{ref}
\bibliographystyle{iclr2025_conference}

\appendix
\section{Appendix}
\subsection{Prompt}
\label{sec:prompt}

\begin{table}[h]
\footnotesize
    \centering
    \begin{tabular}{ |p{13.4cm}| }
    \toprule

Please reason step by step. The final answer without units needs to be placed in \text{\textbackslash boxed\{\}}.\newline
\newline
{\color{blue}\{instruction\}}\newline
 \\
\bottomrule
\end{tabular}
\caption{
Our zero-shot prompt}
\label{our_zero-shot_prompt}
\end{table}

\begin{table}[h]
\footnotesize
    \centering
    \begin{tabular}{ |p{13.4cm}| }
    \toprule

{Review the user's question and the corresponding response using the additive 5-point scoring system described below. Points are accumulated based on the satisfaction of each criterion:\newline
\newline
- Add 1 point if the response is relevant and provides some information related to the user's inquiry, even if it is incomplete or contains some irrelevant content.\newline
- Add another point if the response addresses a substantial portion of the user's question, but does not completely resolve the query or provide a direct answer.\newline
- Award a third point if the response answers the basic elements of the user's question in a useful way, regardless of whether it seems to have been written by an AI Assistant or if it has elements typically found in blogs or search results.\newline
- Grant a fourth point if the response is clearly written from an AI Assistant's perspective, addressing the user's question directly and comprehensively, and is well-organized and helpful, even if there is slight room for improvement in clarity, conciseness or focus.\newline
- Bestow a fifth point for a response that is impeccably tailored to the user's question by an AI Assistant, without extraneous information, reflecting expert knowledge, and demonstrating a high-quality, engaging, and insightful answer.\newline
\newline
\newline
User: {\color{blue}\{question\}}\newline
\newline
$<$response$>${\color{blue}\{response\}}$<$/response$>$\newline
\newline
After examining the user's instruction and the response:\newline
\newline
- Briefly justify your total score, up to 100 words.\newline
- Conclude with the score using the format: ``Score: \text{\textbackslash boxed\{$<$total points$>$\}}"\newline
\newline
Remember to assess from the AI Assistant perspective, utilizing web search knowledge as necessary. To evaluate the response in alignment with this additive scoring model, we'll systematically attribute points based on the outlined criteria.\newline
} \\
\midrule
\end{tabular}
\caption{
LLM-as-a-Judge prompt}
\label{LLM-as-a-Judge_prompt}
\end{table}

\begin{table}[h]
\footnotesize
    \centering
    \begin{tabular}{ |p{13.4cm}| }
    \toprule

{You grade the Solution, verifying correctness step by step. At the end of the Solution verification, when you give your final grade, write it in the form  "Verification: \text{\textbackslash boxed\{X\}}", where X is either Yes or No and placed in the \text{\textbackslash boxed\{\}}.\newline
\newline
**Question**:\newline
{\color{blue}\{question\}}\newline
\newline
**Solution**:\newline
{\color{blue}\{solution\}}\newline
\newline
Let's verify step by step.\newline
}
\\
\midrule
\end{tabular}
\caption{
generative verifier prompt}
\label{gv_prompt}
\end{table}

\FloatBarrier

\subsection{Training Details}
\label{training_detail}
We select the Qwen2.5-7B-Instruct \citep{qwen25} as initial model, which has excellent instruction-following capabilities, making it good for sampling data.
During the solution sampling process, we use our zero-shot prompt and instruct the model to perform step-by-step reasoning, while also requiring the final answer to be enclosed in a LaTeX-style \texttt{\textbackslash boxed\{\}} to better extract and evaluate the model's answer.
In the data construction phase, for each question, model samples 30 solutions with temperature of 1.0. After validating with the gold answer, we pair all correct solutions with incorrect ones and select the top-K pairs with the largest confidence gap (For the MATH task, $K=10$; for the GSM8K task, $K=15$). After collecting the data, we train using batch size of 128 and a learning rate of 5e-7 for 2 epochs with DPO, saving a checkpoint every 50 steps.
Ultimately, we select the best checkpoint from all saved models based on performance on the development set.
As for the coefficient $\beta$ after testing, we find that $\beta=0.3$ yielded the best performance.

\subsection{Examples}
\label{sec:example}

\begin{table}[h]
\footnotesize
\centering
\begin{tabular}{p{13.4cm}}
\toprule
\textbf{Example on MATH}\\ 
\midrule
\textbf{Question}: A standard deck of 52 cards has 13 ranks (Ace, 2, 3, 4, 5, 6, 7, 8, 9, 10, Jack, Queen, King)\newline and 4 suits ($\spadesuit$, $\heartsuit$, $\diamondsuit$, and $\clubsuit$), such that there is exactly one card for any given rank and suit.  Two of the \newline suits ($\spadesuit$ and $\clubsuit$) are black and the other two suits ($\heartsuit$ and $\diamondsuit$) are red.  The deck is randomly arranged. \newline What is the probability that the top card is a $\heartsuit$?
\\
\midrule
\textbf{Gold answer}: There are 13 $\heartsuit$'s and 52 cards total,\newline so the probability that the top card is a $\heartsuit$ is $\dfrac{13}{52} = \boxed{\dfrac14}$.
\\
\midrule
{\textbf{Model's solution}: Let's break down the problem step by step:\newline
\newline
1. There are 52 cards in the deck, of which 13 are $\heartsuit$.\newline
2. The problem asks for the probability that the top card is a $\heartsuit$, which means we only care about the first card.\newline
3. There are 52 possible outcomes (drawing a card from the top of the deck).\newline
4. Since the top card can be any of the 52 cards, the number of favorable outcomes (drawing a $\heartsuit$) is 13.\newline
5. Therefore, the probability can be calculated as:\newline
\newline
P(top card is $\heartsuit$) = Number of favorable outcomes / Number of total outcomes\newline
= 13/52\newline
\newline
\textbf{The final answer without units is:} $\boxed{{\color{red}\frac{1}{4}}}$\newline
\newline
} \\
\midrule
{\textbf{Confidence}: {\color{blue}0.917}}
\newline \\
\midrule
\end{tabular}
\caption{
Confidence Example}
\label{confidence_example}
\end{table}

\begin{table}[ht]
\footnotesize
    \centering
    \begin{tabular}{p{13.4cm}}
    \toprule

\textbf{Example of RewardMATH} \\ \midrule
{\textbf{Question}:$441+2(21)(19)+361=x$. Solve for $x$.
}\\
\midrule
{\textbf{{\color{blue}Correct solution}}:\newline
I want to solve for x in the equation $441+2(21)(19)+361=x$.\newline
I can simplify the equation by first calculating $2(21)(19)$, which is equal to $798$.\newline
Then I can rewrite the equation as $441+798+361=x$. I notice that $441=21^2$ and $361=19^2$. So the equation becomes $21^2+2(21)(19)+19^2=x$.\newline
This is just the binomial expansion of $(21+19)^2$, which equals $40^2$.
Therefore, the solution for x is $\boxed{1600}$.
}\\
\midrule
{\textbf{{\color{red}Wrong solution 1}}:\newline
I want to solve for x in the equation $441+2(21)(19)+361=x$.\newline
I can simplify the equation by first calculating $2(21)(19)$, which is equal to $800$.\newline
Adding these values, $441 + 800 + 361 = 1602$. Therefore, the value of $x$ is $\boxed{1602}$.
}\\
\midrule
{\textbf{{\color{red}Wrong solution 2}}:\newline
Calculate $441+2(21)(19)+361$. $441+2(21)(19)+361 = 441+798+361 = 1190$. Therefore, the value of $x$ is \boxed{1190}.
}\\
\midrule
{\textbf{{\color{red}Wrong solution 3}}:\newline
The expression can be simplified as $441+2(21)(19)+361=x$ $441+2\cdot 21 \cdot 19+361=x$, $441+792+361=x$, $1594=x$ Therefore, the value of $x$ is \boxed{1594}.
}\\
\midrule
{\textbf{{\color{red}Wrong solution 4}}:\newline
We start by simplifying the expression on the left side of the equation.\newline
First, calculate \(2(21)(19)=818\) Then, add \(441 + 818 + 361 = 1620\). Therefore, \(x = \boxed{1620}\)
}\\
\midrule
{\textbf{{\color{red}Wrong solution 5}}:\newline
This is a simple algebraic equation. We follow the order of operations to evaluate the left-hand side of the equation and find the value of $x$.\newline
$441+2(21)(19)+361$, $441+2(399)+361$, because $2 \times 21 \times 19 = 399$ $441+798+361$, because $2 \times 399 = 798$, $1500$, because $441 + 798 + 361 = 1500$\newline
Therefore, $x = 1500$.
}\\
\midrule
{\textbf{{\color{red}Wrong solution 6}}:\newline
Start with the given equation: $441+2(21)(19)+361=x$ Expand the equation:\newline
$441+2(399)+361=x$ Simplify the equation: $1141+x=441+1198$\newline
Combine like terms: 1141+x=1639, Subtract 1141 from both sides: x=1639-1141\newline
Simplify the right-hand side: x=$\boxed{498}$
}\\
\midrule
{\textbf{{\color{red}Wrong solution 7}}:\newline
Calculate the product of 21 and 19. $441+411+361=x$. Combine like terms. $872+361=x$. Add like terms. $1233=x$. Subtract 1233 from both sides. $x=0$.
}\\
\midrule
{\textbf{{\color{red}Wrong solution 8}}:\newline
To solve for $x$, I need to simplify the expression on the left side of the equation.\newline
I notice that there are several factors of 2 in the middle term, so I can use the distributive property to rewrite it as $400+380+361$.\newline
This makes the expression $441+400+380+361$. Now I can add these numbers from left to right, starting with $441+400$.\newline
This gives me $841+380+361$. Next, I add $841+380$ to get $1221+361$. Finally, I add $1221+361$ to get $1582$. So, $x=1582$.
}\\
\midrule
{\textbf{{\color{red}Wrong solution 9}}:\newline
Using the order of operations, we first evaluate $2(21)(19)$ to get $2(378)=756$.\newline
Then, we add $441$, $756$, and $361$ to get $441+756+361=1558$. So, $x=\boxed{1558}$.
}\\
\midrule
\end{tabular}
\caption{
Confidence Example}
\label{rewardmath_example}
\end{table}

\FloatBarrier

\subsection{Derivation}
\label{app::Derivation}
For the given model $\pi$, We define the probability that the solution generated by model $\pi$ is correct as $P_{\pi}(\hat{s}=\text{\small True})$, and the probability that the corresponding answer is the gold answer as $P_{\pi}(\hat{a}=a)$. Thus, we have:
\setcounter{equation}{0}
\begin{equation}
    P_{\pi}(v(\hat{s}, q)=1) = P(\hat{a}=a) \pi(\hat{s}|q)
\end{equation}

For a given question $q$, the probability of the model answering correctly is:

\begin{align}
\label{eq:eq1}
    \mathbb{E}_{\hat{s}\sim \pi}[P_{\pi}(v(\hat{s}, q)=1)] = \mathbb{E}_{\hat{a}\sim \pi}[P(\hat{a}=a) \pi(\hat{s}|q)] = \mathbb{E}_{\hat{a}\sim \pi}[ \mathbb{E}_{\hat{r}\sim \pi} [P(\hat{a}=a)\pi(\hat{a}|\hat{r}, q)] ]
\end{align}
Then, for a given question $q$, for all possible correct solutions \( \hat{s}=<\hat{r}, \hat{a}> \in S^+ \), we have:
\begin{gather*}
    \mathbb{E}_{<\hat{r}, \hat{a}> \in S^+}[\pi_{2}(\hat{a}|\hat{r}, q) - \pi_1(\hat{a}|\hat{r}, q)] \\
    =\mathbb{E}_{<\hat{r}, \hat{a}> \in S^+}[P(\hat{a}=a)\pi_{2}(\hat{a}|\hat{r}, q) - P(\hat{a}=a)\pi_1(\hat{a}|\hat{r}, q)] \\
    =\mathbb{E}_{\hat{a}}[ \mathbb{E}_{\hat{r}} [P(\hat{a}=a)\pi_{2}(\hat{a}|\hat{r}, q) - P(\hat{a}=a)\pi_{1}(\hat{a}|\hat{r}, q) ] ] \\
    =\mathbb{E}_{\hat{a}}[ \mathbb{E}_{\hat{r}} [P(\hat{a}=a)\pi_{2}(\hat{a}|\hat{r}, q) ]] - \mathbb{E}_{\hat{a}}[ \mathbb{E}_{\hat{r}} [ P(\hat{a}=a)\pi_{1}(\hat{a}|\hat{r}, q) ] ] \\
    =\mathbb{E}_{\hat{s}_2 \in S}[P(v(\hat{s}_2, q)=1))] - \mathbb{E}_{\hat{s}_1 \in S}[P(v(\hat{s}_1, q)=1))] \\
    >0 \tag{3}
\end{gather*} 
The above derivation is based on Eq. \ref{eq:eq1} and the fact that 
\begin{align}
    P(v(\hat{s}, q)=1) = 1, \quad\forall \hat{s} \in S^{+}, \notag \\
    P(v(\hat{s}, q)=1) = 0, \quad\forall \hat{s} \in S^{-} \tag{4}
\end{align} 

Similarly, for all incorrect solutions \( \hat{s}=<\hat{r}, \hat{a}> \), we have:
\begin{align}
    \mathbb{E}_{<\hat{r}, \hat{a}> \in S}[\pi_{2}(\hat{a}|\hat{r}, q) - \pi_1(\hat{a}|\hat{r}, q)]&<0 \tag{5}
\end{align} 

\end{document}